\documentclass[sigconf]{acmart}

\usepackage{todonotes}
\usepackage[ruled]{algorithm2e} 
\usepackage{array}
\usepackage{paralist}

\definecolor{Violet}{HTML}{7F00FF}

\AtBeginDocument{%
  \providecommand\BibTeX{{%
    \normalfont B\kern-0.5em{\scshape i\kern-0.25em b}\kern-0.8em\TeX}}}
    
\setcopyright{acmcopyright}
\copyrightyear{2021}
\acmYear{2021}
\acmDOI{10.1145/1122445.1122456}

\begin{document}

\title{ReGroup: Recursive Neural Networks for Hierarchical Grouping of Vector Graphic Primitives}

\author{Sumit Chaturvedi}
\affiliation{
  \institution{Independent}
  \city{NOIDA}
  \country{India}
}
\email{sumit.chaturvedi@gmail.com}

\author{Michal Lukáč}
\affiliation{%
  \institution{Adobe Research}
  \city{San José}
  \country{USA}
}
\email{lukac@adobe.com}

\author{Siddhartha Chaudhuri}
\affiliation{%
  \institution{Adobe Research}
  \city{Bangalore}
  \country{India}
}
\email{sidch@adobe.com}

\renewcommand{\shortauthors}{Sumit, et al.}

\begin{abstract}
Selection functionality is as fundamental to vector graphics as it is for raster data. But vector selection is quite different: instead of pixel-level labeling, we make a binary decision to include or exclude each vector primitive. In the absence of intelligible metadata, this becomes a perceptual grouping problem. These have previously relied on heuristics derived from empirical principles like Gestalt Theory, but since these are ill-defined and subjective, they often result in ambiguity.
Here we take a data-centric approach to the problem. By exploiting the recursive nature of perceptual grouping, we interpret the task as constructing a hierarchy over the primitives of a vector graphic, which is amenable to learning with recursive neural networks with few human annotations.
We verify this by building a dataset of these hierarchies on which we train a hierarchical grouping network. We then demonstrate how this can underpin a prototype selection tool.
\end{abstract}

%
%



\begin{teaserfigure} 
  \includegraphics[width=\textwidth]{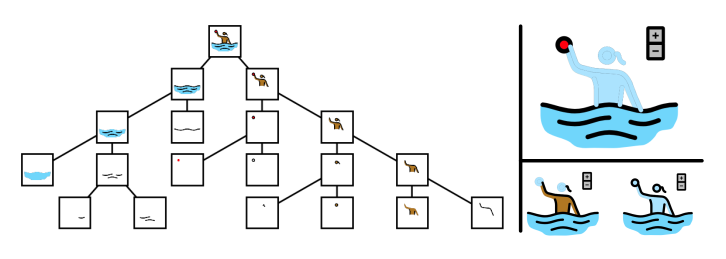}
  \caption{We parse a 2D vector graphics image (root node of the tree on the left) into a perceptually organized hierarchy of graphical elements, and use it to power selection tools. We collect a dataset of such hierarchies and train a \textit{neural network} that learns to group by recursively merging elements. One of our selection tools (\textit{top right}) works by selecting an initial path (the swimmer's face in this example) and using the \textit{plus} button to traverse to ancestors of the path. The expanded selection is highlighted in light blue. Competing state-of-the-art approaches, Fisher et. al.~\cite{fisher2021automatic} (\textit{inset bottom left}) and Suggero~\cite{10.1145/2512349.2512801} (\textit{inset bottom right}), fail to grow the selection correctly, grouping head and ball in one case and omitting all outlines in the other. This is because handwritten heuristics alone are not robust indicators of grouping affinity: our data-driven method learns to avoid such errors.}
  \label{fig:teaser}
\end{teaserfigure}

\maketitle

\section{Introduction}

Vector graphics are the de facto standard for wide classes of diagrams, illustrations and other resolution-independent visuals. They are typically expressed as unions of {\em paths}: open or closed curves with various geometric and color attributes which constitute atomic elements of the drawings. Editing a semantically meaningful component -- such as the head of a human figure or the roof of a house, each of which may comprise several paths -- requires that the component first be {\em selected}. While careful designers can organize paths into meaningful groups to make selection easier, large quantities of graphics ``in the wild'' lack much if any such organization, making selection a tedious and finicky process~\cite{10.1145/2512349.2512801, 10.1145/2366145.2366161}.

Vector editing programs, such as Adobe Illustrator and Inkscape, try to make selecting groups of paths in unorganized graphics easier with tools such as rectangular and lasso selections, in addition to the regular one-click-per-path mode. However, a meaningful component might be too close to other paths and/or non-convex, to make such tools inefficient. ``Smart'' selection tools that are now common for raster images, such as selecting foreground objects or human figures~\cite{barrett2002object}, are much less common in vector-based software. A significant reason for the discrepancy is that deep neural networks, which power modern smart selection, are developed primarily for the raster domain. Nevertheless, some prior research has explored advanced selection mechanisms, such as Lasso and Harpoon~\cite{10.1145/2047196.2047275}, which interpret designers' intentions from imprecise selection strokes. These methods either rely on handwritten heuristics based on {\em Gestalt principles}~\cite{lawsofseeing,wertheimer1938laws} for grouping paths based on perceptual relationships, or ignore perception altogether and analyze textual annotations in the graphic's DOM~\cite{dhanuka2020automatic}.

In this paper, we focus on the common case of {\em perceptual grouping} in the absence of reliable textual annotations. Gestalt principles, while easy for humans to understand, present several practical difficulties for developing automated grouping heuristics. We identify {\bf two key challenges}. First, they are often too ambiguous to robustly {\em operationalize} by hand. For example, the seemingly clear {\em Principle of Proximity} yields many competing options: should we measure proximity based on the distance between path bounding boxes, or between path centroids, or between all path points (average or Hausdorff), or something completely different? Each such choice will fail on a long tail of real-world cases. Second, the heuristics explored in prior work are largely local, operating on pairs of paths at a time and ignoring global context. This actually violates the overarching Gestalt idea of {\em Pragnanz}, according to which visual relations are best understood in the light of the overall harmony of the figure. However, a compact mathematical formulation of Pragnanz is unknown, and unlikely to exist. Hence, our ability to write context-aware grouping rules by hand is extremely limited.

To address these challenges, we propose a {\bf novel, data-driven model} for automatically inferring a {\em grouping hierarchy} for a vector graphic. By training a statistical model on a dataset of human-annotated groupings, to implicitly capture notions of perceptual similarity between vector paths at various scales, we avoid the difficult and error-prone task of explicitly specifying heuristic grouping rules. By forcing the model to encode not just the path(s) of interest, but also their surrounding context from the whole image, we try to ensure a degree of Pragnanz.  By deriving a complete tree whose leaves are atomic paths and internal nodes are selectable groups, we obtain an organizational structure that humans can easily understand and manipulate, which fits standard models of perception and cognition~\cite{serre2014hierarchical}, which can be precomputed in advance, and which enables several applications.

Our method trains a neural network to predict the perceptual affinity between any two disjoint subsets of paths of a graphic. Specifically, the network maps each such subset to a unit vector in a high-dimensional embedding space, where the cosine similarity between two vectors acts as a proxy for their grouping affinity. To infer the hierarchy, we first embed all individual paths (the leaves of the hierarchy). Then, we group the two nearest neighbors in embedding space into a single subtree with a newly-created common parent as the root. The embedding of the union of the leaf paths replaces the individual child embeddings. This process is repeated until a single tree spanning the entire graphic remains. Note that in each step of this hierarchical agglomerative clustering, a new cluster representative (the root embedding of the merged subtree) is computed using a (fixed) neural module, instead of the standard choice of some handwritten statistic such as the mean. Hence, this is a variant of a {\em recursive neural network (RvNN)}~\cite{socher2011parsing} (with one caveat, discussed in Section~\ref{sec:inference}). While recursive models have been explored in the context of text, raster image, and 3D shape parsing~\cite{socher2011parsing,socher2013recursive,li2017grass}, we have to handle the very distinctive grouping cues in stylized vector graphics, where the elements include long, thin, curved and non-convex paths, placed length-to-length, and are not easily separable by approximately convex partitioning. We refine this basic strategy by adding containment constraints~\cite{fisher2021automatic}, a useful and intuitive prior that further regularizes the model.

We show that our method is significantly better at approximating human perceptual grouping than a range of prior baselines including the current state-of-the-art, using a variety of evaluation metrics. We justify our design decisions with ablation studies. We also demonstrate various prototype applications that leverage our inferred hierarchies, such as slider, toggle, scribble, and language-based ``smart'' selection tools.

To summarize, our contributions are as follows:
\begin{itemize}
    \item We argue for the tree data structure as an expressive and concise way to organize vector graphic elements (Section \ref{section:tree}).
    \item We collect a dataset of graphics with perceptual grouping hierarchies (Section \ref{section:data}).
    \item We develop a new neural network model, trained on this dataset, that maps subsets of paths to a common embedding space, where spatial similarity reflects grouping affinity. We perform hierarchical agglomerative clustering in this space to build the tree for a given graphic, using the network as a data-driven embedding function for each new cluster (Section \ref{section:model}).
    \item We rigorously test the model against prior art using quantitative metrics (Sections \ref{section:eval-metrics}, \ref{section:quant} and \ref{section:ablations}), and show that it improves performance across the board.
    \item We demonstrate applications that leverage the inferred hierarchies for easy selection (Section \ref{section:apps}).
\end{itemize}

\section{Related Work} \label{section:relwork}


\paragraph{Perceptually Guided Selection Tools} have two components: perceptual modeling and interactivity. Design decisions for the two components are guided by the interaction medium and the types of graphics. For example, raster-based electronic ink devices~\cite{10.1145/192426.192494}, vector drawings on large interactive white boards~\cite{10.1145/2512349.2512801, 10.1145/2807442.2807455}, sketched drawings~\cite{noris2012smart}, and vector graphics~\cite{10.1145/2366145.2366161, fisher2021automatic, gogia2016insight}. These design decisions cover three categories: feature selection for perceptual similarity, assimilation of features into data structures, and user interactions built on top of them. 
Proximity, orientation and scale based features are common across the works discussed here. Additionally, PerSketch~\cite{10.1145/192426.192494} detects closure, corners and t-junctions. Suggero~\cite{10.1145/2512349.2512801}, Insight~\cite{gogia2016insight} and Fisher et. al.~\cite{fisher2021automatic} measure shape and color similarity. Suggero detects end-point connectivity, a useful feature, given how people use large white boards. Fisher et al. pay close attention to stroke features such as stroke width, line join and dash arrays. Lazy Selection~\cite{10.1145/2366145.2366161} and cLuster~\cite{10.1145/2807442.2807455} measure shape compactness. Notably, Smart Scribbles~\cite{noris2012smart} measures the time of stroke creation. This information, while not always available (we do not assume access to it), is a strong indicator of similarity. 

The pairwise features between graphical elements are condensed into a suitable data structure. Such a data structure should be multi-scale, have a notion of grouping and be expressive enough to deal with ambiguities arising in visual perception. As discussed in Section \ref{section:tree}, trees are an obvious candidate and indeed most of the works discussed here use them. Suggero, Insight, cLuster, Lazy Selection and Fisher et. al. build a single tree or a collection of them. Since it is not \textit{a priori} clear how different features should be combined, a collection of trees helps maintain different perspectives. Suggero, Insight and Lazy Selection build trees by Hierarchical Agglomerative Clustering (HAC)~\cite {mullner2011modern}. cLuster and Fisher et. al. modify HAC. cLuster learns a small set of weights from a dataset of segmented white boards to drive HAC. Fisher et. al. extend HAC to respect containment relations. PerSketch maintains an object lattice. Atomic elements are combined by a rule-based system. Different nodes share atomic elements allowing for multiple sensible interpretations. Lastly, Smart Scribbles maintains a segmentation of the sketch which is refined by the user. The segmentation is inferred from the minimum $k$-cut of the relationship graph.

User interactions are built on top of these data structures. PerSketch, Lazy Selection and Smart Scribbles are scribble based tools. The scribble is compared with candidate nodes in the data structure for curve similarity. Lazy Selection also uses scribble speed to rank candidate nodes. Suggero and cLuster use Harpoon~\cite{10.1145/2047196.2047275}, where selections can be made by cutting across the selection targets. Suggero places a menu with suggestions for further selections. cLuster automatically adds selections based on the initial selection. Fisher et. al. visualize the tree in a panel adjacent to the graphic. Selections can be made by directly clicking on the vertices of the tree. Insight makes selections guided by a sequence of clicks from the user.

In contrast, our features are {\em learned} from a dataset of tree annotations. Section \ref{section:quant} supports our thesis that data-driven features are more faithful to human experience than those discussed above. We use the perceptual model to recursively infer trees for test graphics, which power a scribble-based selection tool, and an ancestor-traversing slider tool.



\paragraph{Learning to Group Discrete Visual Objects.} Grouping means different things in different contexts. For example, when interpreted as semantic or instance segmentation~\cite{yang2020sketchgcn, girshick2015fast, he2017mask,wang2018finegrained}, it is natural to view it as a {\em classification} problem. In the context of object detection~\cite{he2017mask, girshick2015fast, luo2020learning}, it can be viewed as a {\em regression} problem. Regression is also used to approximate a matrix encoding pairwise grouping relations between graphical elements~\cite{li2018universal}. Grammar-based visual parsing~\cite{girshick2011grammar,10.1145/1944846.1944851,martinovic2013earley} view the problem as {\em conditional sampling} from a probabilistic context-free grammar. Socher et al.~\cite{socher2011parsing,socher2013recursive} introduced {\em recursive neural networks} (RvNNs), which train a fixed network module to parse a set of primitives (e.g. words or image fragments) into trees by embedding the primitives into a latent space, repeatedly merging near neighbors in this space, and recursively using the network to update the embeddings for the new merged nodes. RvNNs can be thought of as ``soft'' grammars, where the grammar structure is never explicitly represented. 3D part-based modeling methods like GRASS~\cite{li2017grass} and StructureNet~\cite{mo2019structurenet} leverage RvNNs. We are also inspired by recursive nets, but our method is tailored for 2D vector graphics parsing, and makes different design choices because of the substantial domain difference. Grouping is also viewed as a {\em metric learning} problem: to model perceptual similarity~\cite{10.1145/3130800.3130841} and to enforce global grouping constraints~\cite{li2018universal}. Lastly, it is viewed as a {\em sub-task} in another task: GRASS~\cite{li2017grass} and Li et. al.~\cite{li2018universal} group elements as part of a generative model while Tagger~\cite{greff2016tagger} uses grouping to solve an auxiliary denoising task. 

Since acquiring grouping datasets is hard, approaches have been developed to generalize to unseen categories. For example, some RvNNs~\cite{li2017grass,li2018universal} incorporate grouping as part of a Variational Autoencoder (VAE). Sampling from a manifold of plausible data points ensures that the models see a wide variety of examples during training. Luo et. al.~\cite{luo2020learning} specify a data-driven reward function and use reinforcement learning techniques such as policy gradient and experience replay to explore many trajectories. Lun et. al.~\cite{10.1145/3130800.3130841} and SketchGNN~\cite{yang2020sketchgcn} use creative data augmentation strategies.

Typically, methods can't scale with increasing numbers of graphical elements~\cite{yang2020sketchgcn}. It is hard to enforce pairwise and global grouping constraints simultaneously~\cite{li2018universal}. Another common problem is that of scale invariance. This is addressed by data augmentation strategies~\cite{girshick2015fast}, or image pyramids~\cite{he2015spatial}, or network architecture design~\cite{lin2017feature}. Alternatively, it is addressed using a local coordinate system~\cite{li2018universal}.

In this work, we build a recursive neural model to hierarchically group vector graphics elements. Unlike GRASS or StructureNet, we address the grouping problem head on, using a metric learning framework to directly learn perceptual similarity between graphical elements. We use data augmentation strategies to generalize to unseen examples. Our method suffers as the complexity of the graphic increases: however, the notion of containment graphs, borrowed from Fisher et. al.~\cite{fisher2021automatic}, mitigates this problem. We achieve scale invariance by factoring the representation of graphical element into its location, provided by a bounding box, and its appearance, by rasterizing the graphical element in its bounding box.

\begin{figure*}[t]
    \centering
    \includegraphics[width=\linewidth]{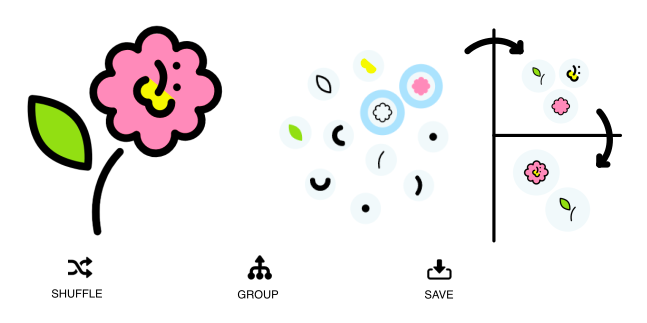}
    \caption{\textit{Annotation Interface}: We show the graphic and its composing paths are side by side. The annotator selects paths by clicking on them directly in the graphic or by clicking on the individual bubbles containing paths. For example, the stroke and fill of the flower petals is selected (highlighted in \textit{blue}). The group button is pressed to combine the chosen paths. The selected bubbles disappear and a new bubble representing the group is added. The annotator continues till only one bubble remains.}
    \label{fig:interface}
\end{figure*}

\section{Trees for Perceptual Organization} \label{section:tree}

\subsection{Notation}

By \textit{trees} we refer to \textit{rooted trees} or \textit{hierarchies}. We setup some notation to ease the upcoming discussion. A tree, $T$ consists of a vertex set $V(T)$ and an edge set $E(T)$. A tree implicitly encodes a partial order among the vertex set. This is the ancestor-descendant relationship between vertices. Let $v_1 \leq v_2$ denote that $v_1$ is a descendant of $v_2$. Further, let $v_1 < v_2$ denote $v_1 \leq v_2 \land v_1 \neq v_2$. Refer to $LCA(x, y)$ as the least common ancestor i.e. the deepest vertex that is also an ancestor of two vertices $x, y \in V$ in $T$. Let $Leaves(T)$ be the leaves of the tree and let $T(x)$ denote the subtree of $T$ rooted at vertex $x$.

\subsection{Alternative Data Structures}

We contrast trees with three alternatives: Graphs, $n$-ary Graphs and single level segmentations.  

Graphics can be organized by recording all pairwise relationships between paths. However, obtaining this data from annotators is expensive. $O(n^2)$ queries need to be made for a single graphic. Moreover, we can't articulate relationships between more than two paths at a time.

$n$-ary Graphs~\cite{mo2019structurenet} are hierarchies with edges between sibling vertices. They are more query-efficient than graphs and are an ideal candidate. As they are a strict generalization of trees, they are more expressive. However, it becomes harder to extract this information from lay people. 

There is substantial prior work on segmentation, both for sketches~\cite{yang2020sketchgcn, 8626530} and images~\cite{10.1145/3130800.3130841}. An algorithm designer has to explicitly choose what a group represents. Is the visual entity revealed by the group a noun, such as an eye, table, pen etc? Or is it merely elements having a common property, such as color, shape, size etc? A tree is better suited to the purpose since it can simultaneously record both high-level, semantic, and low-level, geometry-based relationships.

\subsection{Argument for Trees}

Trees naturally extend clusters and are easier to collect data for (Section \ref{section:data}) than Graphs and \textit{n-ary} Graphs. They have other useful properties for the purpose of perceptual organization. 

First, they can model the strength of the relationship between two percepts. In a tree, $T$, with two vertices, $a$ and $b$, $LCA(a, b)$ denotes the level at which the two percepts, represented by $a$ and $b$, are unified to form a distinct perceptual entity for an observer. The distance between the two vertices in the tree,
\begin{equation}
    Tdist(a, b) = Depth(a) + Depth(b) - 2 Depth(LCA(a,b)),
\end{equation}
is inversely proportional to the strength of their relationship. We use this property to learn an embedding to compare different percepts (Section \ref{section:model}). 

Second, they are efficient for selection. We hypothesize that any \textit{reasonable} selection of $N$ paths can be well approximated by selecting $O(log N)$ vertices in a \textit{well} organized tree. The set of $N$ paths can be written as a union of sets of sizes $2^{\lceil log_2 N \rceil}, 2^{\lceil log_2 N \rceil - 1}, \cdots, 1$. We posit that a vertex in a \textit{well} organized tree can approximate each subset. In other words, only a few vertices would have to be chosen at each level of the tree.

Finally, trees have been a mainstay in describing our visual system~\cite{serre2014hierarchical} and while they only serve as an approximation, it is likely that they endow inductive biases which reduce sample complexity. In fact, people have successfully trained low capacity recursive architectures for 3D shape generation and abstraction~\cite{li2017grass, mo2019structurenet}.

\section{VGTrees Dataset} 
\label{section:data}

We asked annotators to ``assemble'' each graphic hierarchically, starting from an unorganized collection of the paths in the graphic. This notion naturally maps artificial objects, such as chairs, tables, airplanes~\cite{li2017grass, mo2019structurenet, yu2019partnet, wang2011symmetry} and indoor scenes~\cite{li2019grains}, but also generalises to other content well enough that it was useful to convey the task to the annotators.

We also gave annotators additional instructions to help ensure annotated hierarchies map well to usable semantics. We told them to create small-sized groups at a time, typically containing 2-5 subgroups, and we asked that they be able to rationalize their grouping decisions. We gave concrete examples of the latter, including, but not limited to: homogeneity in color or purpose, similarity in shape, and proximity. We also showed them a video where one of the authors performed the task. 

\begin{table}
    \centering
    \begin{tabular}{c|c}
        \# (graphic, tree) pairs & $1508$ \\
        Min paths & 2 \\
        Max paths & 103 \\
        Avg paths & 13.85 \\
        Avg maximum depth in tree & 4.95 \\
        Avg branching factor & 2.32
    \end{tabular}
    \caption{\textit{Dataset statistics at a glance}}
    \label{tab:stats}
\end{table}

\begin{table}
    \centering
    \begin{tabular}{@{}rcccrrr@{}}
        \toprule
        & \# Graphics &$CTED \downarrow$ & \phantom{abc} & \multicolumn{3}{c}{$FMI \uparrow$} \\
        \cmidrule{5-7}
        & & & & $d = 3$ & $d = 2$ & $d = 1$ \\ \midrule
        A v. A & 47 & 0.270 & & 0.128 & 0.504 & 0.863 \\
        A v. C & 233 & 0.317 & & 0.197 & 0.482 & 0.805 \\
        \\
        Overall & 280 & 0.309 & & 0.187 & 0.485 & 0.815 \\
        \bottomrule
    \end{tabular}
    \caption{We ran two studies to check for consistency among different annotators. We compared annotations by two of the authors, \textit{A v. A}. We also compared annotations by one of the authors with crowd sourced ones, \textit{A v. C.}. Each metric is in range $[0, 1]$. Constrained Tree Edit Distance ({\em lower} is better), Fowlkes-Mallows Index ({\em higher} is better). Both show that there is consensus among annotators. In Section \ref{section:quant}, we find that there is more consensus among human annotators, than between humans and algorithms.}
    \label{table:hvh}
\end{table}

We launched the annotation interface (Figure \ref{fig:interface}) as a web application, using graphics sourced from the publicly available OpenMoji dataset~\cite{openMoji}. Table \ref{tab:stats} shows relevant statistics of the dataset. 

The majority of organizations ($\sim 85\%$) were created by the authors. To ensure that this did not cause the dataset to be biased due to annotators being subject experts or prejudiced towards their own methods, we ran two experiments. In the first one, two of the authors each organized 47 of the same graphics. In the second experiment, we crowd-sourced 233 organizations and compared them with our own. We evaluated the agreement between organization using Constrained Tree Edit Distance (CTED)~\cite{zhang1996constrained} and the Fowlkes-Mallows Index (FMI)~\cite{doi:10.1080/01621459.1983.10478008}. In Table \ref{table:hvh}, we find that while consistency between authors (\textit{A v. A}) is better, as is to be expected, there is consistency between an author and crowd sourced annotations too (\textit{A v. C}). In fact, according to CTED and FMI ($d = 1$), there is more consistency between them than between our best model and human annotations. 

\section{Methodology}
\label{section:model}

\subsection{Overview}

Our aim is to organize paths in a graphic into a tree in a way that makes it easy for humans to interact with it. We need to, recursively, decide pairs of paths to group. As the process goes on, grouped paths become candidates for further grouping. 

We use the dataset (Section \ref{section:data}) to learn a function $\phi: V(T) \xrightarrow{} S^{64}$ that maps vertices in a tree to embeddings on a unit hypersphere. We enforce a monotonous mapping between grouping affinity and the cosine similarity of vertex embeddings. 

The function $\phi$ can be used to make grouping decisions by a simple greedy strategy (Algorithm \ref{algo:greedyTree}). Alternatively, we can design other inference strategies that are more suited for certain types of graphics. For example, it is often required that containment relationships be satisfied in the organization. Doing so is a straightforward extension of the scheme that we present, as we describe below.

\subsection{Contrastive Learning for Grouping Affinity}

\begin{figure}
    \centering
    \includegraphics[width=0.9\linewidth]{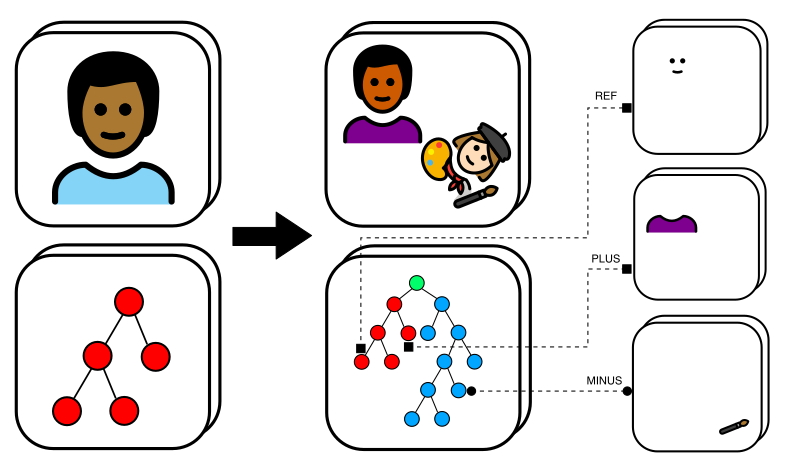}
    \caption{Each data point is a (graphic, tree) pair (\textit{left}). These points go through a data augmentation pipeline (\textit{center}). In this example, another data point is sampled and placed on the canvas. The path fills of the original graphic are perturbed in HSV space. The trees of the original data point (in red) and the new one (in blue) are combined using a new root vertex (in green). Three vertices are sampled (\textit{right}). The REF and the PLUS vertices are closer to each other than to the MINUS vertex in the tree. Grouping affinity between vertices is learned via a contrastive loss.}
    \label{fig:batch-construction}
\end{figure}

We use the contrastive learning framework~\cite{schroff2015facenet, oord2018representation, wang2020understanding} to learn $\phi$, parameterized by a neural network. A batch of triplets is constructed (Figure \ref{fig:batch-construction}) with each triplet containing a reference $v$, a positive $v_{+}$ and a negative $v_{-}$ vertex. The reference and positive vertices have a higher grouping affinity than with the minus vertex. Using a modification of the contrastive loss, InfoNCE~\cite{oord2018representation}, we enforce that this holds true in the embedding space as well.
\begin{equation}
    \mathcal{L} = -\frac{e^{\phi(v) \cdot \phi(v_{+})/\tau}}{e^{\phi(v) \cdot \phi(v_{+})/\tau} + e^{\phi(v) \cdot \phi(v_{-})/\tau}}
\end{equation}
We sample triplets on the basis of distance between vertices in the tree. Three vertices, $v_1, v_2, v_3$ are chosen at random from $V(T)$ and a permutation of $\{1, 2, 3\}$, $(i, j, k)$ is found such that:
\begin{equation} \label{equation:triplet-cond}
    TDist(v_i, v_j) < TDist(v_i, v_k) \wedge TDist(v_i, v_j) < TDist(v_j, v_k)
\end{equation}
Then $v_i$, $v_j$ and $v_k$ become the reference, positive and the negative vertices respectively. We found that adding the second proposition in the conjunction above was crucial in learning a good embedding (Section \ref{section:ablations}). 

In early experiments we used a max margin loss for embeddings in euclidean space and found that training was unstable: our gradient clipping threshold was hit very often. In contrast, this behaviour wasn't observed for the current setup. This has been observed before~\cite{wang2020understanding}. 

\subsection{ReGroup Model}

We now describe how we parameterize the embedding function $\phi$. The model has three components: \textit{location encoder}, \textit{appearance encoder} and \textit{context encoder}. All three process vertices in the tree.

Each graphic is placed in a square canvas. Each vertex, $v \in V(T)$ in the corresponding tree represents a set of paths: $Leaves(T(v))$. We compute the axis oriented bounding boxes for each path. These are normalized to fit in a unit square. The extremities of this unit square represent those of the square canvas. 

The \textit{location encoder} is an MLP with two hidden layers. For a particular vertex, the tightest bounding box covering all its paths is computed and fed to the location encoder as input. The bounding box conveys information on the size and location of the vertex. Our experiments suggest that the \textit{location encoder}, on its own, is surprisingly good at modeling grouping affinity.

The \textit{appearance encoder} is an ImageNet pretrained ResNet18~\cite{he2016deep}. Its input is a raster of all paths in the vertex, composited on a square canvas. The square canvas covers the bounding boxes of the paths and includes some padding. The \textit{fc} layer of the ResNet is modified so that instead of producing the logits for ImageNet classification, it produces a small dimensional vector. 

Together, the \textit{location encoder} and the \textit{appearance encoder} convey the vertex' visual attributes in a resolution independent manner. In order to incorporate contextual information of the vertex in the presence of the rest of the graphic, we add a \textit{context encoder}.  

The \textit{context encoder} is based on the Fast R-CNN modification proposed by Mask R-CNN~\cite{girshick2015fast, he2017mask}. This module is another ImageNet pretrained ResNet18 equipped with the RoIAlign layer. Its input is a raster of the whole graphic and the vertex bounding box. The ResNet18 produces a feature map from the raster. The RoIAlign layer pools the features from the map in the region specified by the vertex bounding box. While these features are biased towards the region of the graphic containing the vertex, their receptive field is the whole graphic. In this sense, they encode how the vertex appears in the context of the whole graphic. Further, we make the same modification to $fc$ of this module as that for the \textit{appearance encoder.}

All three modules described above output a $64D$ vector. These vectors are concatenated and processed by another MLP with two hidden layers to produce an embedding on $S^{64}$. The outputs of the individual modules are layer normalized~\cite{ba2016layer} so that training progresses independently of the difference in initialization of the three modules and to force downstream layers to use all three of them. 

\subsection{Data Augmentation}

By simple XML DOM manipulation, we can create new data points that help enforce desirable invariants. A nice consequence of working with the DOM is that the augmentations are much faster than ones in the raster domain. We list the five augmentations that form a part of our pipeline. 

\begin{enumerate}
    \item \textit{Rotational Transformation}: The graphic is rotated around its center.
    \item \textit{No Fill Transformation}: We set all fill colours to empty to obtain line drawing.
    \item \textit{Stroke Width and Opacity Jitter}: We make strokes thicker or thinner. We also perturb the fill opacities.
    \item \textit{HSV Jitter}: We transform the path fills in HSV space.
    \item \textit{Graphic Combination}: We combine two graphics
    on the same canvas by placing them in non overlapping regions.
\end{enumerate}

Grouping affinities, arguably, don't change under any of these transformations. Figure \ref{fig:batch-construction} shows an example of applying the \textit{Rotational Transformation}, \textit{HSV Jitter} and \textit{Graphic Combination} to the dataset. We use these augmentations in a probabilistic program. As a result, we generate a wide variety of training examples, each a product of zero or more of the augmentations described above. 

\begin{algorithm}[t!]
    \SetAlgoLined
    \SetKwInput{KwData}{Input}
    \KwData{$\phi$, $N$ Paths in Graphic}
    \KwResult{Tree $T$}
    $subtrees$ = [$path_1, path_2, \cdots, path_N$]\;
    \While{length(subtrees) > 1}{
        $subtreePairs$ = $combinations(subtrees, 2)$\;
        $maxSimilarity$ = $-\infty$\;
        $bestPair$ = $undefined$\;
        \For{$pair$ $in$ $subtreePairs$}{
            $(subtree_1, subtree_2)$ = $pair$\;
            $e_1$ = $\phi(subtree_1.root)$\;
            $e_2$ = $\phi(subtree_2.root)$\;
            $sim$ = $e_1 \cdot e_2$\;
            \If{$sim < maxSimilarity$}{
                $maxSimilarity$ = $sim$\;
                $bestPair$ = $pair$\;
            }
        }
        $(subtree_1, subtree_2)$ = $bestPair$\;
        $subtrees.remove(subtree_1)$\;
        $subtrees.remove(subtree_2)$\;
        $subtrees.add(bestPair)$\;
    }
    $T$ = $subtrees.first$\;
    \caption{Greedy Tree Inference}
    \label{algo:greedyTree}
\end{algorithm}

\begin{figure}
    \centering
    \includegraphics[width=\linewidth]{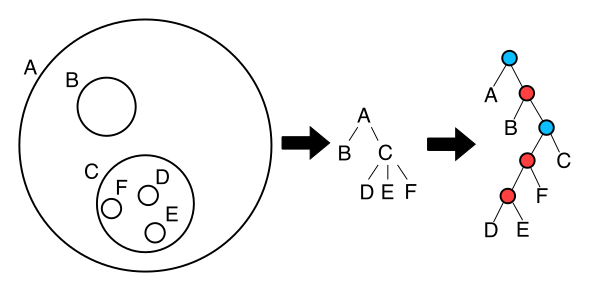}
    \caption{Containment guided tree inference. Given detected containment relationships between paths $A$-$E$ -- $C$ contains $D,E,F$ and $A$ contains all the other paths (left) -- we follow Fisher et al.~\cite{fisher2021automatic} to construct an initial coarse tree based on just these relationships (middle), and then expand each high-degree parent-children substructure to its final deeper form using our ReGroup neural grouping algorithm (right). The resulting structure respects all the containment constraints, while also reflecting the finer organization within each containment group.}
    \label{fig:containmentguided}
\end{figure}

\subsection{Tree Inference}
\label{sec:inference}

We present a simple $O(N^3)$ algorithm (Algorithm \ref{algo:greedyTree}), with $N$ being the number of paths in the graphic, for inferring an organizing hierarchy. Note that since our learned affinity function is cosine similarity in the embedding space, we can leverage efficient nearest neighbor search techniques~\cite{guo2020anisotropic} to accelerate inference in complex graphics. This is one of the principal reasons behind our design decision to target a standard inner-product metric, instead of having the network directly output a scalar affinity score for two paths. 

We start with a set of trivial \textit{subtrees}, each comprising just one node representing one of the $N$ individual paths. At each iteration, the two subtrees with the highest cosine similarity are chosen and merged via a newly-created common parent. The chosen pair is removed from the set and the combined subtree is added: the size of the set reduces by $1$. After $N - 1$ steps, the complete hierarchy is obtained.

This setup is identical to hierarchical agglomerative clustering (HAC), with one crucial difference. Standard HAC algorithms provide only rudimentary means to directly compare internal vertices of the tree. The embedding of an internal vertex is approximated as some statistic (e.g. the mean) of the leaf vertex embeddings. Instead, we use a more powerful and accurate alternative: using $\phi$ to obtain embeddings for internal vertices as well. Hence, we embed each recursively obtained path grouping using the same fixed neural network module, making this a variant of recursive neural networks~\cite{socher2011parsing}. The main difference from standard RvNNs is that the embedding of a parent node is computed by the neural net from the union of the leaf paths, not from the embeddings of its immediate children: computing the union is trivial, and we were able to make this work better in our experiments. It is not a ``deep'' model in the same way as a standard RvNN: however, since the network determines the topology of the tree, the embedding of an internal node is indeed influenced by the embeddings of its descendants.

We can extend this algorithm using the idea of containment groups from Fisher et. al.~\cite{fisher2021automatic}. In many circumstances, containment is a strong indicator of grouping affinity. The algorithm proceeds by constructing a containment tree where parent paths contain child paths. (In general, it is possible for a path to be contained by more than one path. We heuristically choose only one path from among all containing paths as the parent.) Once the containment tree is constructed, each set of sibling paths is hierarchically grouped using the neural net-guided greedy strategy. The resulting subtree is then combined with the containing parent path to replace the containment-based parent-child substructure (Figure \ref{fig:containmentguided}). 

\subsection{Implementation Details}

We trained ReGroup for $28$ epochs. Each epoch had $25600$ triplets batched into sets of $32$. For validation we used $2560$ triplets. While the triplets sampled for training were different for each epoch, the ones for validation were fixed. 

We used Adam~\cite{kingma2014adam} with an initial learning rate of $0.0002$ which was decayed by $0.1$ every $7$ epochs. The temperature hyperparameter, $\tau$, was set to $0.1$. 

The rasters for both the \textit{appearance encoder} and the \textit{context encoder} were $256$ by $256$ pixels. The \textit{conv1}, \textit{bn1}, \textit{layer1} and \textit{layer2} layers of the two ResNet18 were frozen. These are the module names as per the implementation in TorchVision. All the layers that weren't ImageNet-pretrained were initialized using Kaiming Initialization~\cite{he2015delving}.

The graphics were rasterized using a GPU rasterizer, \textit{Pathfinder}. Using this helped us avoid saving and loading rasters to/from disk. Training took $7$ hours on a single Nvidia Tesla T4 GPU.  

\section{Evaluation Metrics} \label{section:eval-metrics}

In this section, we describe metrics that we chose to compare inferred trees against the ground truth.

\subsection{Constrained Tree Edit Distance}

Tree Edit Distances~\cite{BILLE2005217} give us a \textit{global} measure for comparing two trees, $T_1$ and $T_2$. We briefly describe them under the framework of \textit{Tai Mappings}~\cite{kuboyama2007matching}.   

A \textit{Tai Mapping}, $TM$ between $V(T_1)$ and $V(T_2)$, is a bijective mapping with the constraint that for any $(v_1, w_1), (v_2, w_2) \in TM$:
\begin{equation}
    v_1 \leq v_2 \iff w_1 \leq w_2
\end{equation}
$TM$ respects the \textit{ancestor} relationship in the two trees. For a $TM$, let $UM$ denote the set of vertices from both trees that are unmatched. Given a cost of \textit{relabeling}, $r(a, b)$ and a cost of \textit{deletion}, $d(a)$, the cost of $TM$ is defined as: %
\begin{equation}
    cost(TM) = \sum_{(v, w) \in TM} r(v, w) + \sum_{v \in UM} d(v)
\end{equation}
A desirable constraint in our problem setting is that two vertices have non overlapping paths in $T_1$, then the same holds for the corresponding mapped vertices in $T_2$. This is achieved by the \textit{Constrained Tai Mapping}~\cite{zhang1996constrained}. This is a \textit{Tai Mapping} that satisfies: for any $(v_1, w_1), (v_2, w_2), (v_3, w_3) \in CTM$: 
\begin{equation}
    v_3 < LCA(v_1, v_2) \iff w_3 < LCA(w_1, w_2)
\end{equation}
We set delete and relabel costs to be: 
\begin{equation}
    \begin{split}
        d(v) & = 1\\
        r(v, w) & = \frac{Leaves(T_1(v)) \oplus Leaves(T_2(w))}{Leaves(T_1(v)) \cup Leaves(T_2(w))}
    \end{split}
\end{equation}
The minimum cost \textit{Constrained Tai Mapping} gives us the Constrained Tree Edit Distance (CTED). When we report CTED, we make sure to normalize it by the total number of nodes in both trees so that it is in the range $[0, 1]$.

\subsection{Fowlkes-Mallows Index}

\begin{figure}
    \centering
    \includegraphics[width=0.8\linewidth]{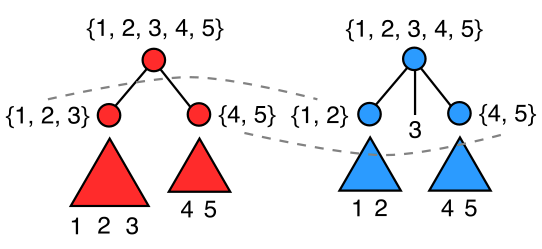}
    \caption{\textit{Fowlkes-Mallows Index} is calculated by aligning two clusterings and measuring their \textit{precision} and \textit{recall}.}
    \label{fig:fmi}
\end{figure}

In addition to tree edit distances, we found it useful to track the $FMI$~\cite{doi:10.1080/01621459.1983.10478008}. It indicates how well the inferred tree, $T_1$ and the ground truth tree, $T_2$, match at a coarse level. For example, as in Figure \ref{fig:fmi}, $T_1$ and $T_2$ are cut at depth $d = 1$ to obtain two clusterings over the paths i.e. the leaves of the trees. For a particular correspondence of clusters, say, $\{1, 2, 3\} \leftrightarrow \{1, 2\}$ and $\{4, 5\} \leftrightarrow \{4, 5\}$, we categorize the pairs as follows:

\begin{enumerate}
    \item \textit{True Positives}: Number of pairs of paths that are in the same cluster on both sides of the mapping.
    \item \textit{False Positives}: Number of pairs of paths that belong to the same cluster on the right hand side of the mapping but not on the left. 
    \item \textit{False Negatives}: Number of pairs of paths that belong to the same cluster on the left hand side but not on the right.
\end{enumerate}

Based on these definitions, we define \textit{precision} and \textit{recall}. The maximum geometric mean of precision and recall, measured over all possible mappings, is the $FMI$ at that level. 

\section{Evaluation} \label{section:quant}

In this section, we report comparisons between our method and the prior state-of-the-art: Suggero~\cite{10.1145/2512349.2512801} and the very recent method of Fisher et al.~\cite{fisher2021automatic}. We evaluate two versions of our method. The first is \textit{ReGroup} where we explicitly enforce containment constraints as part of greedy tree inference. We also evaluate \textit{ReGroup-NoCG}, where we drop the containment constraints. All comparisons, here, are done in a data-driven manner.

We conducted two tests. For the first one, we used different train-val-test splits of VGTrees (Section \ref{section:data}). For the second, we scraped partially organized graphics from the internet.

For the first test, we split VGTrees into 900 (graphic, tree) pairs for training, 200 for validation, and 200 for testing. We ensured that none of the graphics seen during training were included in the validation or test sets. The CTED and FMI scores (Table \ref{table:scores}) for our method are significantly better than for Suggero and Fisher et. al.

For the second test, we scraped novel graphics from \url{Publicdomainvectors.org}. A good number of these graphics, encoded as SVGs~\cite{svgSpecification}, had \textit{group} tags that indicated a grouping of a collection of graphical elements. We filtered for graphics containing three or more of such tags. Around $60\%$ of these graphics contained Inkscape metadata. This led us to believe that groups therein were created by humans, perhaps, for the purpose of editing. We manually inspected hundreds of these groups and found that they were largely consistent with our understanding of perceptual organization. 

Since these graphics are only partially organized, we tried to find out whether the groups in the graphic are approximated by some vertex in our inferred tree. This score, referred to as \textit{Node Overlap}, for a group $G$ and tree $T$, is: 
\begin{equation}
    NO(G, T) = max_{v \in V(T)} IoU(Leaves(T(v)), G)
\end{equation}
We evaluated these scores for 2 subsets of graphics, of size 200 each. We compare 3 grouping methods as in the first test. The scores (Table \ref{table:scores}) are averaged over the 2 subsets and 3 methods.

We found that explicitly enforcing containment relationships was important for achieving the best score on \textit{Node Overlap}: ReGroup improves upon the next best score of Fisher et al. However, this was not strictly true for CTED and FMI scores on VGTrees. One possible explanation is that annotators don't always annotate consistently with respect to these relations, and the purely data-driven pipeline (ReGroup-NoCG) converges to fit these annotations (and hence also the characteristics of the VGTrees test set) more closely.

\begin{table*}[t!]
    \centering
    \begin{tabular}{@{}rccccccc@{}}
        \toprule
        & $NO \uparrow$ & $CTED \downarrow$ & \phantom{abc} & \multicolumn{3}{c}{$FMI \uparrow$} \\
        \cmidrule{5-7}
        & & & & $d = 3$ & $d = 2$ & $d = 1$ \\ \midrule
        Suggero & $0.698 \pm 0.02$ & $0.448 \pm 0.02$ & & $0.193 \pm 0.02$ & $0.385 \pm 0.03$ & $0.660 \pm 0.01$\\
        Fisher et. al. & $0.802 \pm 0.00$ & $0.420 \pm 0.01$ & & $0.239 \pm 0.02$ & $0.432 \pm 0.04$ & $0.691 \pm 0.04$ \\
        \\
        ReGroup-NoCG & $0.747 \pm 0.01$ & $\mathbf{0.343 \pm 0.01}$ & & $0.288 \pm 0.01$ & $\mathbf{0.518 \pm 0.03}$ & $\mathbf{0.779 \pm 0.02}$ \\
        ReGroup & $\mathbf{0.815 \pm 0.01}$ & $0.387 \pm 0.01$ & & $\mathbf{0.296 \pm 0.03}$ & $0.504 \pm 0.04$ & $0.744 \pm 0.03$\\
        \bottomrule
    \end{tabular}
    \caption{\textit{Comparison of our method with prior heuristics-based perceptual grouping methods, using different metrics}: Node Overlap ({\em higher} is better), Constrained Tree Edit Distance ({\em lower} is better), Fowlkes-Mallows Index ({\em higher} is better). Best values in bold. The error bars are 1 standard deviation with 6 trials for NO and 3 for the rest.}
    \label{table:scores}
\end{table*}

\begin{table*}[t!]
    \centering
    \begin{tabular}{@{}rccccccc@{}}
        \toprule
        & $NO \uparrow$ & $CTED \downarrow$ & \phantom{abc} & \multicolumn{3}{c}{$FMI \uparrow$} \\
        \cmidrule{5-7}
        & & & & $d = 3$ & $d = 2$ & $d = 1$ \\ \midrule
        RG-NoCG & $0.747 \pm 0.01$ & ${0.343 \pm 0.01}$ & & $0.288 \pm 0.01$ & $\mathbf{0.518 \pm 0.03}$ & $\mathbf{0.779 \pm 0.02}$ \\
        RG & ${0.815 \pm 0.01}$ & $0.387 \pm 0.01$ & & $\mathbf{0.296 \pm 0.03}$ & $0.504 \pm 0.04$ & $0.744 \pm 0.03$\\ \midrule
        RG - DA & $0.805 \pm 0.01$ & $0.373 \pm 0.01$ & & $0.264 \pm 0.01$ & $0.467 \pm 0.03$ & $0.699 \pm 0.03$\\
        RG-NoCG - DA & $0.709 \pm 0.01$ & $\mathbf{0.335 \pm 0.01}$ & & $0.274 \pm 0.01$ & $0.494 \pm 0.01$ & $0.766 \pm 0.01$\\
        RG + AS & $0.809 \pm 0.01$ & $0.390 \pm 0.01$ & & $0.258 \pm 0.01$ & $0.482 \pm 0.02$ & $0.704 \pm 0.01$\\
        RG-NoCG + AS & $0.730 \pm 0.01$ & $0.364 \pm 0.01$ & & $0.234 \pm 0.01$ & $0.457 \pm 0.01$ & $0.704 \pm 0.01$\\
        RG - L & $0.799 \pm 0.01$ & $0.397 \pm 0.01$ & & $0.284 \pm 0.04$ & $0.492 \pm 0.04$ & $0.723 \pm 0.04$\\
        RG-NoCG - L & $0.701 \pm 0.02$& $0.363 \pm 0.01$ & & $0.250 \pm 0.01$ & $0.480 \pm 0.02$ & $0.748 \pm 0.02$\\
        RG - A & $\mathbf{0.816 \pm 0.01}$ & $0.387 \pm 0.01$ & & $0.280 \pm 0.03$ & $0.501 \pm 0.04$ & $0.747 \pm 0.04$\\
        RG-NoCG - A & $0.771 \pm 0.01$ & $0.352 \pm 0.01$ & & $0.265 \pm 0.01$ & $0.511 \pm 0.03$ & $0.761 \pm 0.02$\\
        RG - C & $0.804 \pm 0.01$ &  $0.401 \pm 0.01$ & & $0.272 \pm 0.04$ & $0.486 \pm 0.04$ & $0.732 \pm 0.04$\\
        RG-NoCG - C & $0.704 \pm 0.01$ & $0.386 \pm 0.02$ & & $0.225 \pm 0.01$ & $0.423 \pm 0.01$ & $0.706 \pm 0.02$\\
        \bottomrule
    \end{tabular}
    \caption{\textit{Ablations of our method}: Node Overlap ({\em higher} is better), Constrained Tree Edit Distance ({\em lower} is better), Fowlkes-Mallows Index ({\em higher} is better). Best values in bold. Legend: \begin{inparadesc}
    \protect\item[RG] ReGroup;
    \protect\item[DA] Data Augmentation;
    \protect\item[AS] Alternative Sampling;
    \protect\item[L] Location Encoder;
    \protect\item[A] Appearance Encoder;
    \protect\item[C] Context Encoder
    \end{inparadesc}. The error bars are 1 standard deviation with 6 trials for NO and 3 for the rest.}
    \label{table:ablations}
\end{table*}

\section{Ablations} \label{section:ablations}

In our ablation study, we evaluated the efficacy of data augmentation, our triplet sampling strategy, and the effect of each of our three encoders.

For data augmentation, our expectation was that data augmentation would aid generalization. Indeed, across all scenarios, models trained with data augmentation performed better than those without. The comparison \textit{Node Overlap} score for ReGroup-NoCG, with and without data augmentation is particularly striking. Especially without the regularization of containment constraints, data augmentation helps achieve better scores on previously unseen graphics. 

To measure the effect of our triplet sampling strategy, we added a second proposition in Equation \ref{equation:triplet-cond} to further constrain the set of triplets. Specifically, we also enforce that $TDist(v_{+}, v) < TDist(v_{+}, v_{-})$. Removing this constraint is detrimental to training, as indicated by the \textit{RG + AS} and \textit{RG-NoCG + AS} scores in Table \ref{table:ablations}.

Finally, we tested the sensitivity of our results to each individual module in ReGroup. One by one, we reinitialized modules in the trained model to Kaiming Initialization. Any patterns learned were, thus, discarded. On the whole, we found that the contribution of each model was important to achieving good scores across metrics. However, surprisingly, reinitializing the \textit{appearance encoder} seems to improve the \textit{Node Overlap} scores for both variants of our method. Reinitializing the \textit{location encoder} led to the biggest drop in \textit{Node Overlap} scores. 

\begin{figure}[t!]
    \centering
    \includegraphics[width=\linewidth]{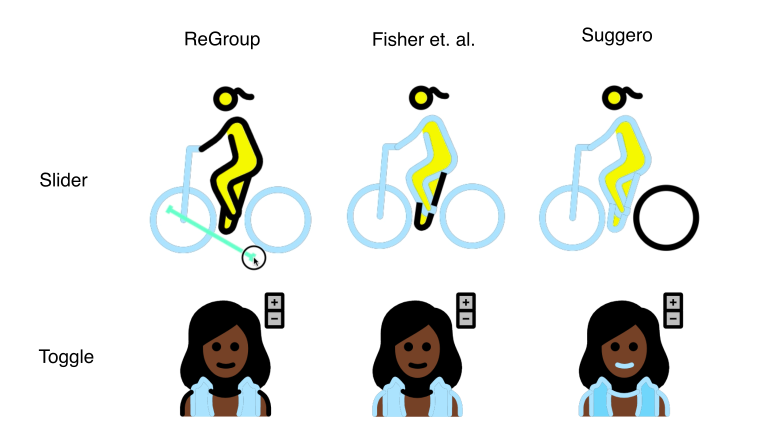}
    \caption{Visualizations of the {\em Slider} and {\em Toggle} selection tools. Please see the supplementary video for a visualization of the {\em Scribble} tool.}
    \label{fig:sliderAndToggle}
\end{figure}

\section{Applications} \label{section:apps}

\subsection{Click-Based Selection Tools}

We made 3 variants of click-based selection tools. We made a \textit{Scribble} tool, using Lazy Selection user interface ~\cite{10.1145/2366145.2366161} to capture user intent, and using the inferred tree to generate meaningful selection suggestions by proposing internal nodes with the highest overlap with the scribbled line. Please see the supplementary video for example interactions with this tool.

We also made two versions of a tool that traverses the ancestor sequence from the selected path to the root of the tree. We call them \textit{Slider} and \textit{Toggle}. 

To use \textit{Slider} (Figure~\ref{fig:sliderAndToggle} (top)), the user clicks on an initial path and drags a slider in any direction. Parent groups are added to the current selection based on the length of the slider. In our initial user study we found that although the tool minimizes cursor travel, it is hard to understand. Users expect the direction of the drag to influence selection. This wasn't the case in our implementation. Moreover, while the drag action was continuous, ancestor traversal was discrete. This made it hard for users to predict what is going to happen next. 

In our final iteration, we developed the \textit{Toggle} tool (Figure~\ref{fig:sliderAndToggle} (bottom)). After an initial click, the toggle bar comes up and users can select ancestors/descendants simply by clicking on \textit{plus}/\textit{minus} button. The interface is one-dimensional and discrete just like the traversal logic.

All three of these tools leverage the precomputed hierarchical structure to enable efficient selection.

\begin{figure}[t!]
    \centering
    \includegraphics[width=\linewidth]{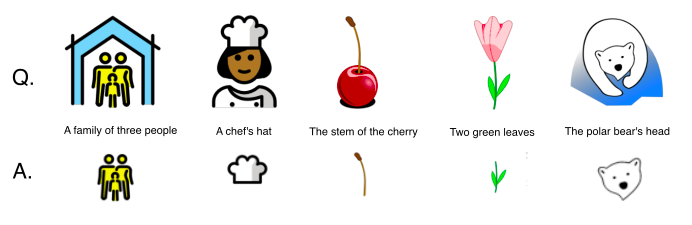}
    \caption{\textit{CLIP for text based selection}: For each graphic (\textit{top row}), and prompt (\textit{middle row}), we find the node in the inferred tree most similar to the prompt using CLIP. The results (\textit{bottom row}) are one of the top 3 most similar nodes for each (graphic, prompt) pair. Examples shown are from both graphics in VGTrees and the public domain. ReGroup infers trees with semantically meaningful nodes, appropriately matching the CLIP prompt.}
    \label{fig:clip}
\end{figure}

\subsection{Text-Based Selection Tool}

We used inferred trees, in combination with CLIP \cite{radford2021learning}, to create a text-based selection tool (Figure \ref{fig:clip}). CLIP is trained to jointly embed text and images in the same space. It is trained on a large dataset of (text, image) pairs and has impressive \textit{zero-shot} capabilities. 

Our prototype asks users for a text prompt. CLIP's text encoder is used to find a feature vector for the prompt. CLIP's image encoder embeds all nodes in the inferred tree in the same feature space. We find which node has the highest cosine similarity with the prompt feature vector. The top 3 most similar nodes are given to the user as suggestions. 

The inferred tree lets us turn an exponential problem, of finding the optimal subset of paths for a prompt, into a linear one, where only the nodes in the tree need to be checked for prompt similarity. This relies on having tree nodes corresponding to most of the semantically meaningful entities of interest. As Figure~\ref{fig:clip} demonstrates, and the Node Overlap metric quantifies, this is frequently the case.

\section{Limitations and Future Work}  \label{section:limitations}

While our method achieves good qualitative and quantitative results on our test cases, its design induces several limitations. These are related to what relationships we can capture, and what hierarchies we can construct.

Because the greedy hierarchical clustering algorithm only considers pairs of nodes, we can only infer binary trees. This means we cannot directly represent e.g. the three blades of a ceiling fan or the spots on a ladybug in a single group: instead, they are grouped into an arbitrary binary tree. Because our node similarity measure is capable of evaluating similarity between leaves as well as groups, it should be possible to extend the greedy algorithm to use this information to add nodes to an existing group instead of creating a new one.

In the process of rendering the graphic for appearance and context encoders, the render order information is discarded as the paths are composited. This could potentially discard important hints on objects to be grouped based on e.g. being close in the render order. Future work could address this perhaps by adding a \textit{position encoding}~\cite{vaswani2017attention} to the embeddings. 

We have in VGTrees, discovered several designs that violate our assumption that every path corresponds to an atomic design element. For instance, we found a graphic of two people standing side by side, where their torsos were represented by a single path. In this case, our model is unable to construct a semantically meaningful hierarchy. However, it is easy for a designer to fix this by splitting up such a compound object, and perhaps future work might discover a way of doing this automatically.

There are multiple possible ways of evaluating containment constraints, and which is the optimal one seems to be situationally dependent. For example, it is not clear how to decide which objects, if any, are contained by an open path. In our implementation, we make the pragmatic decision of connecting the end points of open paths, which is optimal in terms of our evaluation scores, but we can still find examples where this is not the right thing to do.

Moreover, always following containment constraints can give rise to false positives. Consider the example of a figure holding a water bottle in front of their torso. In this case the containment hint is spurious, because it represents the spatial relationship of separate objects rather than there being a single discrete object. But since the bottle is ``contained'' in the torso, it'll be grouped with it before the torso is grouped with the hands. We hope there may be a data-driven solution for classifying these meaningful versus spurious containment relationships.

Our method only extracts what we believe to be semantic groups without attempting to actually recover the semantics by e.g. estimating vertex labels.  As a result, we don't know what a particular group represents. This also ultimately limits our ability to exploit the semantics of the groups to further improve grouping by considering e.g. the logical relationships between objects. Overcoming this is difficult, since the content is often highly stylised or abstract, and not all levels of a hierarchy are necessarily semantically meaningful.

Lastly, our method has not been stress-tested on complex graphics. It would also benefit from access to much larger training datasets. Unlike heuristic-based methods, whose performance is fixed, our data-driven approach has the potential to continuously improve as it is trained on more data.

\section{Conclusion} 

In this paper we present a method to automatically organise vector graphics into hierarchical structures represented as trees, for the purposes of simplifying the manipulation of these graphics. We have proposed a data-driven similarity metric which we trained on the dataset we had gathered, and designed an algorithm to construct a hierarchy from this similarity information.

We evaluated our method against various baselines on a variety of vector data, focusing on the veracity of the induced hierarchy compared to the ground truth. Finally, we demonstrated prototype selection tools which take advantage of the hierarchical structure to facilitate rapid semantic selection.


\bibliographystyle{ACM-Reference-Format}
\bibliography{bibliography}

\end{document}